\documentclass{article}

\usepackage{dblfloatfix}
\usepackage{microtype}
\usepackage{graphicx}
\usepackage{subfigure}
\usepackage{booktabs}
\usepackage{xspace}
\usepackage{amsmath,amsfonts,amssymb}
\usepackage{tikz}
\usepackage{bm}
\usepackage{pdfpages}
\usetikzlibrary{calc}
\usetikzlibrary{decorations.text}
\usetikzlibrary{shapes.geometric}

\usepackage{mathtools}
\DeclarePairedDelimiter\ceil{\lceil}{\rceil}
\def\defeq{\mathrel{\stackrel{\rm def}=}}
\newtheorem{remark}{Remark}

\usepackage{hyperref}
\usepackage{balance}

\definecolor{color1}{HTML}{A8E6CE}
\definecolor{color2}{HTML}{DCEDC2} 
\definecolor{color3}{HTML}{FFD3B5} 
\definecolor{color4}{HTML}{FF8C94} 

\definecolor{color5}{HTML}{dcedc1} 

\newcommand{\nodeclr}{white}
\newcommand{\sigmaclr}{white}
\newcommand{\noderadinline}{0.6ex}
\newcommand{\nodelinewidth}{1.5pt}
\newcommand{\edgelinewidth}{1.5pt}
\newcommand{\drawparam}{black}

\newcommand{\tbasis}{\mbox{T-Basis}\xspace}
\newcommand{\tensor}[1]{\mathbf{#1}}
\newcommand{\tensorcal}[1]{\bm{\mathcal{#1}}}
\newcommand{\scalar}[1]{\mathit{#1}}
\newcommand{\R}{\mathbb{R}}

\newcommand{\nlayers}{L}
\newcommand{\bsize}{B}
\newcommand{\tspace}{\mathbb{TB}}
\newcommand{\rpm}{\raisebox{.2ex}{$\scriptstyle\pm$}}

\usepackage[accepted]{icml2020}

\icmltitlerunning{\tbasis: a Compact Representation for Neural Networks}

\begin{document}

\twocolumn[
\icmltitle{\tbasis: a Compact Representation for Neural Networks}

\icmlsetsymbol{equal}{*}

\begin{icmlauthorlist}
\icmlauthor{Anton Obukhov}{ethz}
\icmlauthor{Maxim Rakhuba}{ethz}
\icmlauthor{Stamatios Georgoulis}{ethz}
\icmlauthor{Menelaos Kanakis}{ethz}
\icmlauthor{Dengxin Dai}{ethz}
\icmlauthor{Luc Van Gool}{ethz,kuleuven}
\end{icmlauthorlist}

\icmlaffiliation{ethz}{ETH Zurich}
\icmlaffiliation{kuleuven}{KU Leuven}

\icmlcorrespondingauthor{Anton Obukhov}{anton.obukhov@vision.ee.ethz.ch}

\icmlkeywords{neural, network, compression, tensor, ring, basis}

\vskip 0.3in
]

\printAffiliationsAndNotice{}

\begin{abstract}
We introduce \tbasis, a novel concept for a compact representation of a set of tensors, each of an arbitrary shape, which is often seen in Neural Networks.
Each of the tensors in the set is modeled using Tensor Rings, though the concept applies to other Tensor Networks.
Owing its name to the T-shape of nodes in diagram notation of Tensor Rings, \tbasis is simply a list of equally shaped three-dimensional tensors, used to represent Tensor Ring nodes.   
Such representation allows us to parameterize the tensor set with a small number of parameters (coefficients of the \tbasis tensors), scaling logarithmically with each tensor's size in the set and linearly with the dimensionality of \tbasis. 
We evaluate the proposed approach on the task of neural network compression and demonstrate that it reaches high compression rates at acceptable performance drops. 
Finally, we analyze memory and operation requirements of the compressed networks and conclude that \tbasis networks are equally well suited for training and inference in resource-constrained environments and usage on the edge devices.
Project website: \href{http://obukhov.ai/tbasis}{obukhov.ai/tbasis}.
\end{abstract}

\section{Introduction}
\label{introduction}

Since the seminal work of Krizhevsky et al.~\yrcite{krizhevsky2012imagenet}, neural networks have become the ``go-to" approach for many research fields, including computer vision~\cite{redmon2016you,he2017mask,chen2018encoder}, medical image analysis~\cite{ronneberger2015u,kamnitsas2017efficient}, natural language processing~\cite{graves2013speech,devlin2018bert}, and so on. This tremendous success can largely be attributed to a combination of deeper architectures, larger datasets, and better processing units. It is safe to say that neural networks have gradually turned into deep networks that contain millions of trainable parameters and consume lots of memory. For example, the ResNet-101 model~\cite{he2016deep} has 44M parameters and requires 171MB of storage, while the VGG-16 model~\cite{simonyan2014very} has 138M parameters and requires 528MB of storage. Most importantly, further advancements in network performance seem to go hand-in-hand with a corresponding increase in the network size.

On the other hand, over the last few years, we have been witnessing a steady transition of this technology to industry. Thus, it is becoming a pressing problem to deploy the best-performing deep networks to all kinds of resource-constrained devices, such as mobile robots, smartphones, wearables, and IoT devices. These systems come with restrictions in terms of runtimes, latency, energy, and memory consumption, which contrasts with the considerations behind the state-of-the-art approaches. 
At the same time, the lottery ticket hypothesis~\cite{frankle2018lottery} and several other works~\cite{denil2013predicting,ba2014deep} suggest strong evidence that modern neural networks are highly redundant, and most of their performance can be attributed to a small fraction of the learned parameters.

Motivated by these observations, \emph{network compression} has been proposed in the literature to arrive at smaller, faster, and more energy-efficient neural networks. 
In general, network compression techniques can be grouped into the following categories: pruning, hashing, quantization, and filter/tensor decomposition -- we provide a detailed discussion of each category in Sec.~\ref{related_work}. In our work, we build upon the tensor decomposition front, and in particular, the Tensor Ring (TR) decomposition~\cite{zhao2016tensor}, due to its high generalization capabilities when compressing convolutional (and fully-connected) layers~\cite{wang2018wide}. Formally, TR decomposes a high-dimensional tensor as a sequence of third-order tensors that are multiplied circularly. 
For example, a $3 \times 3$ convolutional kernel with 16 input and 32 output channels, i.e., a $16 \times 32 \times 3 \times 3$ tensor with $4608$ entries, may admit a parameterization using TR with rank 2, defined by $4$ tensors of the sizes $2 \times 16 \times 2$, $2 \times 32  \times 2$, $2 \times 3 \times 2$, $2 \times 3 \times 2$ with only $4 (16 + 32 + 3 + 3) = 216$ total entries.
However, a common drawback of existing TR approaches on network compression, like TRN~\cite{wang2018wide}, is that they have to estimate an individual TR factorization for each tensor (i.e., layer) in the network.

In this paper, we go beyond this limitation and introduce \tbasis. In a nutshell, instead of factorizing the set of arbitrarily shaped tensors in the neural network, each with a different TR representation and independent parameters, we propose to learn a single basis for the whole network, i.e., the \tbasis, and parameterize the individual TRs with a small number of parameters -- the coefficients of the \tbasis. 

We organize our paper as follows.
In Sec.~\ref{sec:method}, we present the concept of \tbasis (Sec.~\ref{sec:T-basis}) and its application to the compression of convolutional neural networks (Sec.~\ref{sec:tens}).
Sec.~\ref{sec:impl} is devoted to implementation details such as initialization procedure (Sec.~\ref{sec:impl_init}) and complexity estimates (Sec.~\ref{sec:convappl}).
The numerical results are presented in Sec.~\ref{experiments}.

\section{Related Work}
\label{related_work}

\paragraph{Pruning} 
Pruning methods attempt to identify the less important or redundant parameters in a trained neural network and prune them, to reduce the inference time while retaining performance. 
Han et al.~\yrcite{han2015learning,han2016deep} proposed to prune `low impact' neurons, while other approaches focused more on pruning filters/channels~\cite{li2017pruning,gordon2018morphnet,yang2018netadapt}, as the latter leads to more regular kernel shapes~\cite{li2019learning}. 
Structured sparsity constraints~\cite{alvarez2016learning,zhou2016less,wen2016learning,liu2017learning} have typically been employed to achieve the desired level of pruning. They usually enforce channel-wise, shape-wise, or depth-wise sparsity in the neural network. 
Alternatively, He et al.~\yrcite{he2017channel} used Lasso regression-based channel selection and least square reconstruction, and Yu et al.~\yrcite{yu2018nisp} proposed a Neuron Importance Score Propagation algorithm. 
In general, offline pruning suffers from the need to fine-tune the pruned network to mitigate the performance drop, while online pruning requires to properly balance the regularization and task losses, which is not straightforward. 
Moreover, Liu et al.~\yrcite{liu2019rethinking} recently questioned the usefulness of pruning, claiming that similar, or even better, results could be achieved by training an equivalent network from scratch.  

\paragraph{Hashing} In the context of neural networks, hashing can be used to create a low-cost mapping function from network weights into hash buckets that share a single parameter value for multiple network weights. Based on this observation, Chen et al.~\yrcite{chen2015compressing} proposed HashNets that leverage a hashing trick to achieve compression gains. Follow-up works leveraged different techniques (i.e., Bloomier filters) for the weight encoding~\cite{reagen2017weightless} or developed more efficient ways to index the hashing function~\cite{spring2017scalable}. Recently, Eban et al.~\yrcite{eban2019structured} proposed using multiple hashing functions to partition the weights into groups that share some dependency structure, leading to considerable compression gains. However, in hashing methods, once the hashing function is created, it can not easily be adapted, which limits the network's transfer and generalization ability.

\paragraph{Quantization} Another straightforward technique to reduce the size of a neural network is by quantizing its weights. Popular quantization schemes include lower precision~\cite{wu2016quantized,jacob2018quantization,faraone2018syq,nagel2019data}, binary~\cite{courbariaux2015binaryconnect,courbariaux2016binarized,rastegari2016xnor,zhou2016dorefa}, or even ternary~\cite{zhu2016trained} weights. 
Quantization and hashing are orthogonal to other compression techniques, including ours, and can be used in conjunction with them to achieve further compression gains
-- one example is 
~\cite{han2016deep}.

\paragraph{Filter Decomposition} Motivated by Denil et al.~\yrcite{denil2013predicting} who showed the existence of redundancy in the network weights, researchers utilized low-rank matrix decomposition techniques to approximate the original filters.
Denton et al.~\yrcite{denton2014exploiting} investigated different approximations (e.g., monochromatic, bi-clustering) as well as distance metrics for the kernel tensor of a filter, Jaderberg et al.~\yrcite{jaderberg2014speeding} proposed to decompose a $k \times k$ filter into a product of $k \times 1$ and $1 \times k$ filters, while Zhang et al.~\yrcite{zhang2015accelerating} considered the reconstruction error in the activations space. 
The aforementioned low-rank decompositions operate on a per-filter basis (i.e., spatial or channel dimension). However, Peng et al.~\yrcite{peng2018extreme} showed that better compression could be achieved if we exploit the filter group structure of each layer. Li et al.~\yrcite{li2019learning} expanded upon this idea by further learning the basis of filter groups. Overall, filter decomposition is a limited case of tensor decompositions described below, and hence these works can typically achieve only moderate compression. 

\paragraph{Tensor Decomposition} Similarly to filter decomposition, various low-rank parameterizations can be utilized for weight tensors of a convolutional neural network~\cite{su2018tensorial}.
For example, tensor decompositions such as the CP-decomposition~\cite{hitchcock-sum-1927} and the Tucker decomposition~\cite{tucker-factor-1963} (see review~\cite{kolda-review-2009}) have been used for compression purposes~\cite{lebedev2014speeding,kim2015compression}. 
Tensor Networks (TNs) with various network structures, like MPS, PEPS and MERA, have been proposed in the literature -- we refer the reader to~\cite{orus2019tensor,larskres-survey-2013} for an overview. 
One popular structure of TNs for neural network compression is Tensor-Train (TT)~\cite{oseledets2011tensor}, which is equivalent to MPS. 
TT decomposes a tensor into a set of third-order tensors, which resembles a linear ``train'' structure, and has been used for compressing the fully-connected~\cite{novikov2015tensorizing} or convolutional~\cite{garipov2016ultimate} layers of a neural network. 
Another structure of TNs, with arguably better generalization capabilities~\cite{wang2017efficient}, is Tensor-Ring (TR)~\cite{perez2006matrix,khor-qtt-2011,espig-networks-2011,zhao2016tensor}. 
TR has been successfully used to compress both the fully connected and convolutional layers of DNNs~\cite{wang2018wide}. 
The proposed \tbasis extends upon the TR concept, enabling the compact representation of a set of arbitrarily shaped tensors. 

\section{Method}
\label{sec:method}

We set out to design a method for neural network weights compression, which will allow us to utilize weights sharing, and at the same time, have a low-rank representation of layers' weight matrices. Both these requirements are equally important, as controlling the size of the shared weights pool permits compression ratio flexibility, while low-rank parameterizations are efficient in terms of the number of operations during both training and inference phases.

A common drawback of many weight sharing approaches is the non-uniform usage of the shared parameters across the layers. For example, in~\cite{li2019learning} the ``head" of the shared tensor is used in all blocks, while the ``tail" contributes only at the latter stages of CNN. Likewise,~\cite{eban2019structured} exploit a hand-crafted weight co-location policy. With this in mind, our method's core idea is to promote uniform parameter sharing through a basis in the space of tensor ring decompositions, which would require only a few per-layer learned coefficient parameters.

We introduce the notation in Sec.~\ref{preliminaries}, then we formulate \tbasis for a general set of tensors in Sec.~\ref{sec:T-basis}, and for convolutional neural network (CNN) compression in Sec.~\ref{sec:tens}.

\subsection{Preliminaries}
\label{preliminaries}

We use the following notation for terms in equations: 
$\scalar{a}$~-- scalars; 
$\bm{a}$~-- matrices; $\bm{a}_{i j}$ -- element of a matrix $\bm{a}\in\mathbb{R}^{m\times n}$ in the position $(i, j)$, $i\in\{1,\dots,m\}$, $j\in\{1,\dots,n\}$;
$\tensor{A}$ (or $\tensorcal{A}$)~-- tensors; 
$\tensor{A}(i_1, i_2, ..., i_d)$ or $\tensor{A}_{i_1, i_2, ..., i_d}$~-- element of a $d$-dimensional tensor $\tensor{A}\in \mathbb{R}^{N_1\times\dots\times N_d}$ in the position $(i_1, i_2, ..., i_d)\in \mathcal{I}$, where $\mathcal{I} = \mathcal{I}_1\times\dots\times \mathcal{I}_d$, $\mathcal{I}_k = \{1,\dots, N_k\}$.
We also write $d(\tensor{A}) = d$.

We say that a tensor~$\tensor{A}\in\mathbb{R}^{N_1\times\dots\times N_d}$ is represented using the TR decomposition if $\forall (i_1, i_2, ..., i_d)\in \mathcal{I}$
\begin{equation} \label{eq:trdec}
    \tensor{A}(i_1, i_2, ..., i_d) = 
    \sum_{\substack{r_1,\dots,r_d=1, \\ r_{d+1}\equiv r_1}}^{R_1,\dots, R_d} \ 
    \prod_{k=1}^d 
        \tensorcal{C}_k (r_k, i_k, r_{k+1}),
\end{equation}
where $\tensorcal{C}_k\in\mathbb{R}^{R_k \times N_k \times R_{k+1}}$, $R_{d+1}=R_1$ are called \emph{cores}.
We refer to $\mathbf{R} = (R_1,\dots,R_d)$ as the \emph{TR-rank}.
Note that~\eqref{eq:trdec} is called \emph{tensor train} decomposition if $R_1 = 1$.

\begin{figure}
\begin{center}
\resizebox{0.35\textwidth}{!}{
\begin{tikzpicture}
\def \n {5}
\def \radius {1.2cm}
\def \nsize {0.9cm}
\def \legsize {0.3cm}
\def \margin {57.2958*\nsize/\radius/2}
\foreach \s in {1,...,\n}
{
  \node[draw, circle, fill=\nodeclr, minimum size=\nsize, line width=\nodelinewidth,draw=\drawparam] (N-\s) at ({360/\n * (\s - 1)}:\radius) {$\tensorcal{C}_{\s}$};
  \draw[-,line width=\edgelinewidth] ({360/\n * (\s - 1)+\margin}:\radius) 
  arc ({360/\n * (\s - 1)+\margin}:{360/\n * (\s)-\margin}:\radius);
  \node at ({360/\n * (\s - 1.5)}:\radius+7) {$r_{\s}$};
  \draw[-,line width=\edgelinewidth] ({360/\n * (\s - 1)}:{\radius + \nsize/2}) -- ({360/\n * (\s - 1)}:{\radius + \nsize/2 + \legsize});
  \node at ({360/\n * (\s - 1)+\margin/3}:{\radius + \nsize/2 + \legsize/1.3}) {$i_{\s}$};
}
\coordinate (Acent) at (-3*\radius, 0);
\node[draw, circle, fill=\nodeclr, minimum size=\nsize,line width=\nodelinewidth,draw=\drawparam] at (Acent) {$\mathbf{A}$};
\node at (-1.75*\radius, 0) {$=$};
\foreach \s in {1,...,\n}
{
    \coordinate (B) at ($ (Acent) + ({360/\n * (\s - 1)}:{\nsize/2}) $);
    \coordinate (C) at ($ (Acent) + ({360/\n * (\s - 1)}:{\nsize/2 + \legsize}) $);
    \coordinate (cooind) at ($ (Acent) + ({360/\n * (\s - 1)}:{\nsize/2 + \legsize + \legsize/1.5}) $);
  \draw[-,line width=\edgelinewidth] (B) -- (C);
  \node at (cooind)  {$i_{\s}$} ;
}
\end{tikzpicture}
}
\end{center}
\caption{A graphical representation of the TR decomposition of $\mathbf{A}\in\mathbb{R}^{N_1\times\dots\times N_5}$. Node $\tensorcal{C}_k$ and the (three) adjacent edges represent a three-dimensional array with entries $\tensorcal{C}_k(r_k, i_k, r_{k+1})$, $r_{6}=r_1$, $i_k\in \mathcal{I}_k$ $r_k \in \{1,\dots, R_k\}$. The edge between two nodes $\tensorcal{C}_k$ and $\tensorcal{C}_{k+1}$ represents summation along the index~$r_{k+1}$. }
\label{fig:trdec}
\end{figure}
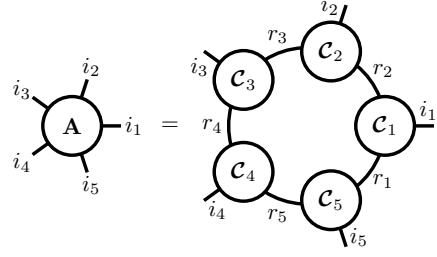

The decomposition~\eqref{eq:trdec} can be conveniently illustrated using tensor diagrams.
Let 
\resizebox{!}{0.02\textwidth}{
\begin{tikzpicture}
\draw[fill=\nodeclr,line width=\nodelinewidth,draw=\drawparam]  circle(0.9*\noderadinline);
\draw[line width=\edgelinewidth] (0:0.9*\noderadinline) -- (0:2.7*\noderadinline);
\draw[line width=\edgelinewidth] (72:0.9*\noderadinline) -- (72:2.7*\noderadinline);
\draw[line width=\edgelinewidth] (144:0.9*\noderadinline) -- (144:2.7*\noderadinline);
\draw[line width=\edgelinewidth] (216:0.9*\noderadinline) -- (216:2.7*\noderadinline);
\draw[line width=\edgelinewidth] (288:0.9*\noderadinline) -- (288:2.7*\noderadinline);
\end{tikzpicture}
}
with $d$ edges denote a $d$-dimensional array, where each edge corresponds to one index~$i_k\in\mathcal{I}_k$ in $(i_1,\dots,i_d)\in\mathcal{I}$.
For example, 
\begin{tikzpicture}
\draw[fill=\nodeclr,line width=\nodelinewidth,draw=\drawparam] circle(\noderadinline);
\draw[line width=\edgelinewidth] (-\noderadinline,0) -- (-3*\noderadinline,0);
\draw[line width=\edgelinewidth] (\noderadinline,0) -- (3*\noderadinline,0);
\end{tikzpicture}
represents a matrix illustrating the fact that each matrix element depends on two indices.
If an edge connects two nodes, there is a summation along the corresponding index.
Therefore, a matrix-vector product can be represented as 
\begin{tikzpicture}
\draw[fill=\nodeclr,line width=\nodelinewidth,draw=\drawparam] circle(\noderadinline);
\draw[line width=\edgelinewidth] (-\noderadinline,0) -- (-3*\noderadinline,0);
\draw[line width=\edgelinewidth] (\noderadinline,0) -- (3*\noderadinline,0);
\draw[fill=\nodeclr,line width=\nodelinewidth,draw=\drawparam] (4*\noderadinline,0) circle(\noderadinline);
\end{tikzpicture}
.
Thus, equality~\eqref{eq:trdec} can be illustrated in Fig.~\ref{fig:trdec} (for $d=5$): on the left-hand side, there is a five-dimensional array $\mathbf{A}$, while on the right-hand side, there are five three-dimensional cores connected in a circular fashion (ring).

\subsection{\tbasis for a Set of Tensors} 
\label{sec:T-basis}

\begin{figure*}[b]
\begin{center}
\resizebox{0.75\textwidth}{!}{
\begin{tikzpicture}[square/.style={regular polygon,regular polygon sides=4}]
\def \n {5}
\def \coretocore {3.5cm}
\def \coretosigma {1.75cm}
\def \nsize {1.3cm}
\def \ssize {0.5cm}
\def \legsize {0.3cm}
\def \margin {57.2958*\nsize/\radius/2}
\node (S1) at (-\coretosigma,0) {};
\draw [rounded corners=3mm, line width=\nodelinewidth,draw=\drawparam] (S1) ++(-1,0) rectangle ++(14.5,1.1);
\node[draw, circle, fill=\nodeclr, minimum size=\nsize, line width=\nodelinewidth,draw=\drawparam] (C1) at (0,0) {${\footnotesize \negmedspace\tensorcal{C}_1^{(\ell)}\!\negthickspace\in\negthickspace\tspace\negmedspace}$};
\node[draw, circle, fill=\nodeclr, minimum size=\nsize, line width=\nodelinewidth,draw=\drawparam] (C2) at (\coretocore,0) {${\footnotesize \negmedspace\tensorcal{C}_2^{(\ell)}\!\negthickspace\in\negthickspace\tspace\negmedspace}$};
\node[draw, circle, fill=\nodeclr, minimum size=\nsize, line width=\nodelinewidth,draw=\drawparam] (C3) at (2*\coretocore,0) {$\dots$};
\node[draw, circle, fill=\nodeclr, minimum size=\nsize, line width=\nodelinewidth,draw=\drawparam] (C4) at (3*\coretocore,0) {${\footnotesize \negmedspace\tensorcal{C}_{d_\ell}^{(\ell)}\!\negthickspace\in\negthickspace\tspace\negmedspace}$};
\node[draw, square, fill=\sigmaclr, minimum size=\ssize, line width=\nodelinewidth,draw=\drawparam] (S1) at (-\coretosigma,0) {${\footnotesize \negmedspace\negmedspace\negmedspace\bm{\rho_1^{(\ell)}}\negmedspace\negmedspace\negmedspace}$};
\node[draw, square, fill=\sigmaclr, minimum size=\ssize, line width=\nodelinewidth,draw=\drawparam] (S2) at (-\coretosigma+\coretocore,0) {${\footnotesize \negmedspace\negmedspace\negmedspace\bm{\rho_2^{(\ell)}}\negmedspace\negmedspace\negmedspace}$};
\node[draw, square, fill=\sigmaclr, minimum size=\ssize, line width=\nodelinewidth,draw=\drawparam] (S3) at (-\coretosigma+2*\coretocore,0) {${\dots}$};
\node[draw, square, fill=\sigmaclr, minimum size=\ssize, line width=\nodelinewidth,draw=\drawparam] (S4) at (-\coretosigma+3*\coretocore,0) {${\footnotesize \negmedspace\negmedspace\negmedspace\bm{\rho_{d_\ell}^{(\ell)}}\negmedspace\negmedspace\negmedspace}$};
\foreach \s in {1,...,4}
{
\draw[-,line width=\edgelinewidth] (S\s) -- (C\s);
\draw[-,line width=\edgelinewidth] (C\s) -- ($ (C\s)+(0,-1.2cm) $);
}
\foreach \s in {1,...,3}
{
\pgfmathtruncatemacro{\sp}{\s + 1}
\draw[-,line width=\edgelinewidth] (C\s) -- (S\sp);
}
\end{tikzpicture}
}
\end{center}
\caption{A graphical representation of $\tensor{A}_\ell$, $\ell\in\{1,\dots,\nlayers\}$ using the \tbasis concept. Each $\tensor{A}_\ell$ is represented using  TR-decompositions with TR-cores belonging to $\tspace$ defined in \eqref{eq:tspace}. Diagonal matrices $\bm{\rho}_k^{(\ell)}\in\R^{R\times R}$ with positive diagonal entries allow for rank adaptation.}
\label{fig:t-basis}
\end{figure*}
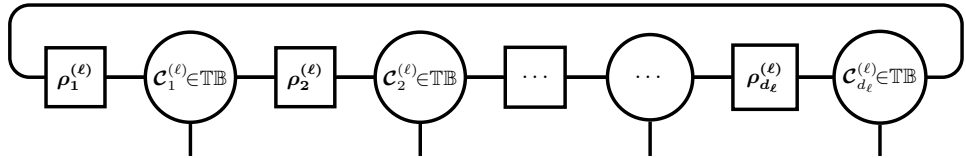

Suppose we are given a set of $\nlayers$ tensors 
\begin{equation}\label{eq:collection}
    \mathbf{A}_\ell \in\mathbb{R}^{N\times\dots \times N},\ \   d(\mathbf{A}_\ell) = d_\ell,\ \ \ell\in\{1,\dots,\nlayers\},
\end{equation} 
each given by the TR decompositions with the cores $\mathbf{C}_k^{(\ell)}$.
We additionally assume that $R_1= \dots = R_d = R$, which is often done when utilizing the TR decomposition.

For the set of tensors defined above, we introduce \emph{\tbasis}: a set of $\bsize$ three-dimensional arrays $\tensorcal{B}_\beta\in\mathbb{R}^{R\times N \times R}$, $\beta \in \{1,\dots,\bsize\}$ such that every core can be represented using this basis.
In other words, we impose the condition that all the TR-cores $\mathbf{C}_k^{(\ell)}$ belong to a \tbasis subspace defined as
\begin{equation}\label{eq:tspace}
    \tspace \defeq \biggl\{\sum_{\beta=1}^{\bsize} \lambda_\beta \tensorcal{B}_\beta \Bigm| \lambda_\beta \in \R  \biggr\} \subseteq \R^{R\times N \times R}.
\end{equation}
As a result, for each $\ell$, there exists a matrix of coefficients $\bm{\alpha}^{(\ell)} \in \mathbb{R}^{d_\ell\times\bsize}$:
\begin{equation}\label{eq:tbasis}
    \tensorcal{C}_k^{(\ell)} = 
        \sum_{\beta = 1}^{\bsize} \bm{\alpha}^{(\ell)}_{k\beta}\, \tensorcal{B}_\beta,
    \quad
    k\in\{1,\dots,d_\ell\}.
\end{equation}
Note that such a basis always exists with $\bsize =  \mathrm{dim}(\R^{R\times N \times R}) = NR^2$.
Nevertheless, a noticeable reduction in storage is possible when $\bsize\ll NR^2$, which is the case in all our experiments.

To introduce additional flexibility to the model, we utilize diagonal matrices~$\bm{\rho}_k^{(\ell)} \in\R^{R\times R}$ termed \emph{rank adaptors}, between every pair of adjacent cores:
\[
    \tensor{A}_\ell (i_1, ..., i_d) = 
    \negthickspace\negthickspace\negthickspace
    \sum_{\substack{r_1,\dots,r_d=1, \\ r_{d+1}\equiv r_1}}^{R,\dots, R} \ 
    \prod_{k=1}^d 
        (\bm{\rho}_k^{(\ell)})_{r_{k}r_{k}} \tensorcal{C}_k^{(\ell)} (r_k, i_k, r_{k+1}).
\]
The purpose of rank adaptors is two-fold. First, we want parameters of the basis and coefficients to be initialized using a normal distribution (Sec.~\ref{sec:impl_init}) and not have them diverged too much from it, which serves the purposes of better training stability and facilitates \tbasis transfer (e.g., taking a pre-trained basis and using it as is to parameterize and train a different network). This goal may not be entirely achievable through the learned \tbasis coefficients. Second, as seen in SVD decomposition of a matrix $\tensor{A}=\tensor{U}\tensor{S}\tensor{V}^\top$, singular values correspond to its Frobenius norm, and their magnitude defines which columns of $\tensor{U}$ and $\tensor{V}$ matrices have a bigger contribution to the product. Thus rank adaptation can be seen as an attempt to promote extra variability in the tensor ring compositions space. Similarly to SVD, we experimented with different constraints on the values of rank adapters, and found out that keeping them non-negative gives the best results, as compared to having them signed, or even not using them at all; see Sec.~\ref{experiments}.
The resulting \tbasis concept with the rank adapters is illustrated in Fig.~\ref{fig:t-basis}.

\paragraph{Storage Requirements}
Let us estimate the storage requirements for the \tbasis compared with TR decomposition.
Assume that $d_1=\dots=d_\nlayers = d$.
To represent all the arrays $\tensor{A}_\ell$, $\ell\in\{1,\dots,\nlayers\}$, one needs to store $L$ coefficient matrices of the sizes $d\times\bsize$, $\nlayers d$ diagonals of rank adapters, and the basis itself.
At the same time, TR decompositions of $\tensor{A}_\ell$ require storing $d\nlayers$ cores of sizes $R\times N \times R$.
Therefore, the storage ratio $\mathsf{r}$ of the two approaches is
\[
\begin{split}
    \mathsf{r} = \frac{\mathrm{mem}_{\text{\tbasis}}}{\mathrm{mem}_{\text{TR}}} &=
        \frac{\nlayers d \bsize+\nlayers dR + \bsize NR^2}{d\nlayers NR^2} = \\
        &=\frac{\bsize}{NR^2} + \frac{1}{NR} + \frac{\bsize}{d\nlayers} = \mathsf{r}_{\bm{\alpha}} + \mathsf{r}_{\bm{\rho}} + \mathsf{r}_{\tensorcal{B}}.
\end{split}
\]
Typical values in the numerical experiments: $\bsize \in \{1,\dots,128\}$, $N\in \{4,9\}$, $R \in \{2,\dots, 32\}$, $\nlayers = \#\text{NN-layers} = \{32,56\}$ in ResNet CIFAR experiments, $d_\ell \leq 10$, leading to $\mathsf{r}^{-1}\leq 200$. 

\begin{remark}
Let $S$ be the number of weights in the largest layer of a CNN.
Then, the total storage of the whole network is at worst $LS$.
Simultaneously, for a fixed \tbasis, storing coefficients and rank adapters requires only $\mathcal{O}(L \log_n S)$.
\end{remark}

Next, we will discuss the procedure of preparing layers of a CNN to be represented with the shapes as in~\eqref{eq:collection}. 

\subsection{\tbasis for CNN Compression} 
\label{sec:tens}

One layer of a convolutional neural network (CNN) maps a tensor $\tensor{X} \in \R^{W\times H \times C^\mathrm{in}}$
to another tensor $\tensor{Y} \in \R^{(W-K+1)\times (H-K+1) \times C^\mathrm{out}}$ with the help of a tensor of weights $\mathbf{W} \in \R^{C^\mathrm{out} \times C^\mathrm{in} \times K \times K}$, such that
\begin{equation}\label{eq:conv}
    \tensor{Y}(w, h, i) =
    \negthickspace\negthickspace
        \sum_{p,q,j=1}^{K, K, C^\mathrm{in}}
        \negthickspace\negthickspace
            \tensor{W}(i,j,p,q)
            \tensor{X}(w\!+\! p-\! 1, h\!+\! q\! - \! 1, j)
\end{equation}
for all $w\in\{1,\dots,W-K+1\}$, $h\in\{1,\dots,H-K+1\}$, and $i\in\{1,\dots,C^\mathrm{out}\}$.
Here $C^{\mathrm{out}}$ and $C^\mathrm{in}$ are the numbers of output and input channels; $K$ is the filter size.

Let us first assume that 
\begin{equation}\label{eq:assumption0}
    C^{\mathrm{in}} = C^{\mathrm{out}} = n^{d}, \quad K = n.
\end{equation}
This specific condition allows us to tensorize (reshape) the tensor of weights $\tensor{W}$ into a $2(d+1)$-dimensional array $\tensor{\widetilde W}$ of size $n\times n\times  \dots \times n$ such that
\begin{equation}\label{eq:tens_v0}
    \tensor{\widetilde W}(i_1,\dots,i_d,j_1,\dots,j_d, p, q) = \tensor{W}(i,j,p,q),
\end{equation}
where $i_k,j_k\in \{1,\dots,n\}$ and are uniquely defined by 
\[
    i = 1 + \sum_{k=1}^{d} (i_k-1) n^{d-k}, \quad j = 1 + \sum_{k=1}^{d} (j_k-1) n^{d-k}.
\]
We will write $i = \overline{i_1\dots i_d}$, $j = \overline{j_1\dots j_d}$ for brevity.
It is known~\cite{oseledets2011tensor} that tensor decompositions of matrices using scan-line multi-indexing (where factors of $j$ follow factors of $i$) often require a larger rank between the last factor of $i$ and the first factor of $j$ than between other pairs.
Therefore, we use permutation of indices:
\begin{equation}\label{eq:tens_v1}
    \tensor{\widetilde{\widetilde W}}({i_1, j_1},\dots,{i_d, j_d}, p, q) = \tensor{W}(i,j,p,q).
\end{equation}
Finally, we do pairwise index merging that reshapes $\tensor{\widetilde{\widetilde W}}$ to a $(d+1)$-dimensional tensor $\tensorcal{W}\in\R^{n^2\times n^2 \times \dots \times n^2}$:
\begin{equation}\label{eq:tens}
    \tensorcal{W}(\overline{i_1 j_1},\dots,\overline{i_d j_d}, \overline{pq\vphantom{j}}) = \tensor{W}(i,j,p,q).
\end{equation}
Thus, we can already represent layers satisfying $C^{\mathrm{in}}_\ell = C^{\mathrm{out}}_\ell = n^{d_\ell}$ and $K = n$
by reshaping tensors of weights $\tensor{W}_\ell$ by analogy with~\eqref{eq:tens} and applying the proposed \tbasis concept (Sec.~\ref{sec:T-basis}) with $N=n^2$.

Nevertheless, for most of the layers, the assumption~\eqref{eq:assumption0} does not hold.
To account for general sizes of the layers, we first select $n\geq K$ and pad each $\tensor{W}_\ell$, $\ell\in\{1,\dots,\nlayers\}$ to an \emph{envelope} tensor $\tensor{W}^n_\ell$ of the shape $n^{d_\ell} \times n^{d_\ell} \times n \times n$ with
\begin{equation} \label{eq:dimension}
    d_\ell = \max\left\{\ceil*{\log_n C^{\mathrm{out}}_\ell}, \ceil*{\log_n C^{\mathrm{in}}_\ell}\right\},
\end{equation}
where $\ceil*{\cdot}$ is the ceiling operator.
After that, we apply the \tbasis parameterization to the set of tensors $\tensorcal{W}_\ell^n$.
Case of $n<K$ can be handled similarly with padding, tensorization, and grouping of factors of convolutional filter dimensions.

We proceed to learn the joint parameterization of all layers through a two-tier parameterization (\tbasis and Tensor Rings), in an end-to-end manner. During the forward pass, we compute per-layer TR cores from the shared basis and per-layer parameters, such as coefficients and rank adapters. Next, we can either contract the TR cores in order to obtain envelope tensors $\tensor{W}_\ell^n$, or we can use these cores to map an input tensor $\tensor{X}$ directly. In both cases, we zero out, crop, or otherwise ignore the padded values.

\section{Implementation}
\label{sec:impl}

Based on the conventions set by the prior art \cite{novikov2015tensorizing,garipov2016ultimate,wang2018wide}, we aim to perform reparameterization of existing Neural Network architectures, such as ResNets \cite{he2016deep}, without altering operator interfaces and data flow. As a by-product, \tbasis layers can be either applied in the low-rank space or decompressed into the full tensor representation of layer weights and applied regularly (see Sec.~\ref{sec:convappl}). 

\subsection{Reparameterization of Neural Networks}
\label{sec:impl_reparam}

We start reparameterization by creating \tbasis tensors $\tensorcal{B}_\beta$, $\beta\in\{1,\dots,\bsize\}$ from hyperparameters $B$, $R$, and $N$. We analyze these hyperparameters in Sec.~\ref{experiments}; however, a few rules of thumb for their choice are as follows: 
\begin{itemize}
    \item $N$ matching the size of spatial factors of convolutional kernels is preferred (9 for networks with dominant $3\times3$ convolutions, 4 for non-convolutional networks);
    \item Linear increase of $R$ leads to a quadratic increase in the size of \tbasis, but only a linear increase in the size of parameterization. We found that increasing the rank gives better results than increasing basis in almost all cases;
    \item Linear increase of $B$ leads to a linear increase of sizes of \tbasis and weights parameterization and should be used when the rank is saturated. For obvious reasons, the size of the basis should not exceed $NR^2$.
\end{itemize}

Next, we traverse the neural network and alter convolutional and linear layers in-place to support \tbasis parameterization. Specifically, for each layer, we derive tensorization, permutation, and factors grouping plans (Eq.~\eqref{eq:tens_v1}, \eqref{eq:tens}, \eqref{eq:dimension}) for weight tensor, and substitute the layer with the one parameterized by $\bm{\alpha}^{(\ell)}$ and $\bm{\rho}^{(\ell)}$. Similarly to the previous works, we keep biases and BatchNorm layers intact, as their total size is negligible. We also make an exception for the very first convolutional layer due to its small size and possible inability to represent proper feature extraction filters when basis or rank sizes are small.

\subsection{Initialization}
\label{sec:impl_init}

Proper weight initialization plays a vital role in the convergence rate and quality of discovered local minima. Following the best practices of weight initialization \cite{he2015delving}, we wish to initialize every element of the weight tensors $\tensor{W}_\ell$ by sampling it as an i.i.d. random variable from $\mathcal{N}(0,\sigma_{\ell}^2)$, where $\sigma_{\ell}$ is determined from the shape of weight tensor.
However, now that each weight tensor is represented with internal coefficients $\bm{\alpha}^{(\ell)}$,~$\bm{\rho}^{(\ell)}_k$, and \tbasis tensors~$\tensorcal{B}_\beta$ shared between all layers, we need a principled approach to initialize these three parameter groups. To simplify derivation, we initialize all rank adapters $\bm{\rho}^{(\ell)}$ with identity matrices. Assuming that the elements of $\tensorcal{B}_\beta$ and $\bm{\alpha}^{(\ell)}$ are i.i.d. random variables sampled from $\mathcal{N}(0, \sigma_{\tensorcal{B}}^2)$ and $\mathcal{N}(0, \sigma_{\bm{\alpha}^{(\ell)}}^2)$ respectively, the variance of the weight tensors elements is given as follows:
\begin{equation}\label{eq:variance_equiv1}
    \mathrm{Var}\left(\tensorcal{C}_k^{(\ell)}(\cdot)\right) = 
    \mathrm{Var}\left(\sum_{\beta = 1}^{\bsize} \bm{\alpha}^{(\ell)}_{k\beta}\, \tensorcal{B}_\beta(\cdot)\negthickspace\right) =  {\bsize} \sigma_{\tensorcal{B}}^2 \sigma_{\bm{\alpha}^{(\ell)}}^2
\end{equation}
and
\begin{equation}\label{eq:variance_equiv2}
\begin{split}
    &\mathrm{Var}\left(\tensor{W}^{(\ell)}(i_1, ..., i_d)\right) = \\
    &=\mathrm{Var}\left({\small\sum_{\substack{r_1,\dots,r_d=1}}^{R,\dots, R}}\ \prod_{k=1}^d \tensorcal{C}_k^{(\ell)} (r_k, i_k, r_{k+1}) \right) =  \\
    &\qquad=R^d \left( {\bsize} \sigma_{\tensorcal{B}}^2 \sigma_{\bm{\alpha}^{(\ell)}}^2 \right)^d =
    \left(BR\ \sigma_{\tensorcal{B}}^2 \sigma_{\bm{\alpha}^{(\ell)}}^2 \right)^d
    \equiv \sigma_{\ell}^2.
\end{split}
\end{equation}
Equations \eqref{eq:variance_equiv1} and \eqref{eq:variance_equiv2} tie up variance of basis tensor elements with variances of coefficients matrices elements across all layers. 
To facilitate \tbasis reuse, we choose $\sigma_{\tensorcal{B}}$ as a function of basis size and rank, and the rest of the variance is accounted in $\sigma_{\bm{\alpha}^{(\ell)}}$:
\begin{equation}\label{eq:variances}
    \sigma_{\tensorcal{B}}^2 = (BR)^{-1},\quad
    \sigma_{\bm{\alpha}^{(\ell)}}^2 = \sigma_{\ell}^{2/d}
\end{equation}
We perform initialization in two steps. First, we initialize the basis and coefficients parameters using~\eqref{eq:variances}. Next, for each layer $\ell$, we compute the variance of its elements and perform variance correction by scaling the matrix $\bm{\alpha}^{(\ell)}$.

\subsection{Convolutional Operator and its Complexity} 
\label{sec:convappl}

For simplicity, we assume in this section that the size of a convolutional layer satisfies assumption~\eqref{eq:assumption0} and denote $C\equiv C^{\mathrm{in}}=C^{\mathrm{out}}$.
Let us discuss how to implement~\eqref{eq:conv} provided that the tensor of weights $\tensor{W}$ is given by the proposed \tbasis approach.
In this case, there are two possibilities: 
(1) assemble the tensor $\tensor{W}$ from its TR cores, then calculate the convolution regularly, or (2) perform convolution directly in the decomposed layer's low-rank space.

\paragraph{Decompression}
In the former case, we need to assemble $\tensorcal{W}\in\R^{n^2\times\dots\times n^2}$~\eqref{eq:tens} with $d(\tensorcal{W}) = d + 1 = \log_n C + 1$~\eqref{eq:dimension} from its TR decomposition.
Let us estimate the complexity of this operation.
The building block of this procedure is the contraction of two neighboring tensors from the tensor network diagram: given $\tensor{A}\in\R^{R\times N_1 \times \dots \times N_{d} \times R}$ and a tensor $\tensor{B}\in\R^{R\times N_{d+1} \times \dots \times N_{D} \times R}$, we compute $(\tensor{A}\times\tensor{B})\in \R^{R\times N_1 \times \dots \times N_{D} \times R}$ introduced as
\[
\begin{split}
(\tensor{A}\times\tensor{B})(\alpha,i_1,\dots,i_D, \beta)& \defeq \\
    \sum_{\gamma=1}^{R} \tensor{A}(\alpha,i_1,\dots&,i_{d}, \gamma)  \tensor{B}(\gamma, i_{d+1},\dots,i_{D}, \beta),
\end{split}
\]
which requires $\mathcal{O}(R^3N_1\dots N_D)$ operations.
To compute~$\tensorcal{W}$, we perform multilevel pairwise contractions $\times$ of the cores.
Namely, we first evaluate $\tensorcal{C}_k\coloneqq \bm{\rho}_k\times\tensorcal{C}_k$, $k\in\{1,\dots,d+1\}$ (here we consider the $R\times R$ diagonal rank adaptor matrices $\bm{\rho}_k$ as $R\times 1 \times R$ tensors to apply the $\times$ operation), which leads to $\mathcal{O}(d R^2 n^2)$ operations due to the diagonal structure of $\bm{\rho}_k$. Then we compute $\tensorcal{C}_{12} = \tensorcal{C}_1\times \tensorcal{C}_2$, $\tensorcal{C}_{34} = \tensorcal{C}_3\times \tensorcal{C}_4$, $\dots$, which requires $\mathcal{O}((d/2) R^3 n^4)$ operations.
On the next level, we compute $\tensorcal{C}_{1234} = \tensorcal{C}_{12}\times \tensorcal{C}_{34}$, $\tensorcal{C}_{5678} = \tensorcal{C}_{56}\times \tensorcal{C}_{78}$, ..., which, in turn, has $\mathcal{O}((d/4) R^3 n^8)$ complexity.
Assuming for simplicity that $d=2^k-1$ for some $k\in\mathbb{N}$ and after repeating the procedure $(k-1)$ times, we obtain two arrays $\tensorcal{C}_{1\dots (d+1)/2}$ and $\tensorcal{C}_{((d+1)/2+1)\dots (d+1)}$ that share the same rank indices. Using them, we can compute $\tensorcal{W}$ in $R^2 n^{2d}$ operations.
Since computing $\tensorcal{C}_{12\dots (d+1)/2}$ and $\tensorcal{C}_{((d+1)/2+1)\dots (d+1)}$ costs $\mathcal{O}(R^3 n^d)$ and since $d = \log_n C$ and we chose $n=K$, the total complexity is $\mathcal{O}(R^3 n^{d+1} + R^2 n^{2d+2}) = \mathcal{O} (R^3 CK + R^2 C^2K^2)$.

\paragraph{Direct Mapping}
Alternatively, we can avoid assembling the full weight tensor and manipulate directly the input tensor $\tensor{X}$~\eqref{eq:conv}, the TR-cores $\tensorcal{C}_k$ and the rank adapter matrices $\bm{\rho}_k$.
Particularly, we need to calculate the sum
\begin{equation}\label{eq:conv_impl}
\begin{split}
 \tensor{Y}(w, h, \overline{i_1\dots i_d}) = 
    \negthickspace\negthickspace
        \sum_{p,q,j=1}^{K, K, C}
        \negthickspace\negthickspace
        \tensorcal{W}(\overline{i_1 j_1},\dots,\overline{i_d j_d}, \overline{pq\vphantom{j}}) 
        \\
            \tensorcal{X}(w\!+\! p-\! 1, h\!+\! q\! - \! 1, j_1,\dots,j_d)
\end{split}
\end{equation}
where we reshaped $\tensor{X}$ of the size $W\times H \times C$ to an array $\tensorcal{X}$ of the size $W\times H \times n \times \dots \times n$ such that
$
   \tensorcal{X}(w, h, j_1,\dots, j_d) = \tensor{X}(w, h, \overline{j_1\dots j_d}). 
$
In~\eqref{eq:conv_impl} we will first sum over the index $j_1$, which is present only in the first core $\tensorcal{C}_1$ of the tensor $\tensorcal{W}$:
\[
\begin{split}
    \tensorcal{Y}_{1}& (w,h, i_1, j_2,\dots, j_d, r_1, r_2) = \\
    &\sum_{j_1=1}^n   \tensorcal{C}_1(r_1,\overline{i_1 j_1}, r_2) \tensorcal{X}(w, h, j_1,\dots, j_d),
\end{split}
\]  
which costs $\mathcal{O}(WHR^2n^{d+1})$ operations.
Next step is to sum over $r_2$ and $j_2$, which are present in $\tensorcal{Y}_{1}$, $\tensorcal{C}_2$ and in $\bm{\rho}_2$:
\[
\begin{split}
    &\tensorcal{Y}_{2} (w,h, i_1, i_2, j_3,\dots, j_d, r_1, r_3) = \\
    &\sum_{j_2,r_2=1}^{n,R}  
    \negthickspace
    (\bm{\rho}_2)_{r_2 r_2}   \tensorcal{C}_2(r_2,\overline{i_2 j_2}, r_3) \tensorcal{Y}_{1} (w,h, i_1,\dots, j_d, r_1, r_2),
\end{split}
\]  
which, in turn, costs $\mathcal{O}(WH R^3 n^{d+1})$.
Similarly, summation over $r_k$ and $j_k$, $k=2,\dots,d$ will result in the same complexities. 
The complexity of the last step ($k=d+1$) is $\mathcal{O}(WH K^2 R^2 n^d)$.
Thus, the total complexity is $\mathcal{O}(WH R^2 n^{d} (dRn + K^2))$, which is equal to $\mathcal{O}(WH R^3 C\log C)$ for fixed $n=K$.

\section{Experiments}
\label{experiments}

In this section, we investigate the limits of neural network compression with the \tbasis parameterization on two tasks~-- image classification, and a smaller part on the semantic image segmentation.

We evaluate model compression as the ratio of the numbers of independent parameters of the compressed model and the baseline. Unless specified otherwise, the number of independent parameters includes compressible parameters, and the incompressible ones not subject to parameterization (as per Sec.~\ref{sec:impl_reparam}) but excludes buffers, such as batchnorm statistics. It is worth reiterating that techniques like parameter quantization or hashing can be seen as complementary techniques and may lead to even better compression.

On top of the conventional compression ratio, we also report the ratio of the number of compressed model parameters excluding basis, to the number of uncompressed parameters. This value gives an idea of the basis--coefficients allocation trade-off and provides an estimate of the number of added parameters in a multi-task environment with the basis shared between otherwise independent task-specific networks.

In all the experiments, we train incompressible parameters with the same optimizer settings as prescribed by the baseline training protocol. For the rest of the learned parameters, we utilize Adam~\cite{adam} optimizer in classification tasks and SGD in semantic segmentation. 
For Adam, we found that 0.003 is an appropriate learning rate for most of the combinations of hyperparameters. 
Larger sizes of basis require lower learning rates in all cases. We perform linear LR warm-up for 2000 steps in all experiments.

The L2 norm of decompressed weights grows unconstrained in all experiments, approaching the limit of the floating point data type. To counter that, we impose an additional regularization term on the norm with weight 3e-4.

We report Top-1 accuracy for classification and mean Intersection-over-Union (mIoU) for segmentation. All reported values are the best metrics obtained over validation sets during the whole course of training; a few experiments report confidence intervals.
All experiments were implemented in PyTorch~\cite{NEURIPS2019_9015} and configured to fit into one conventional GPU with 11 or 16GB of memory. 

Due to GPUs being more efficient with large monolith tensor operations, we chose to perform decompression of \tbasis parameterization at every training step, followed by a regular convolution, as explained in Sec.~\ref{sec:convappl}. At inference, decompression is required only once upon initialization.

\begin{figure}[t]
\centering
\includegraphics[trim={0 0 0 2cm},clip,width=1.05\linewidth]{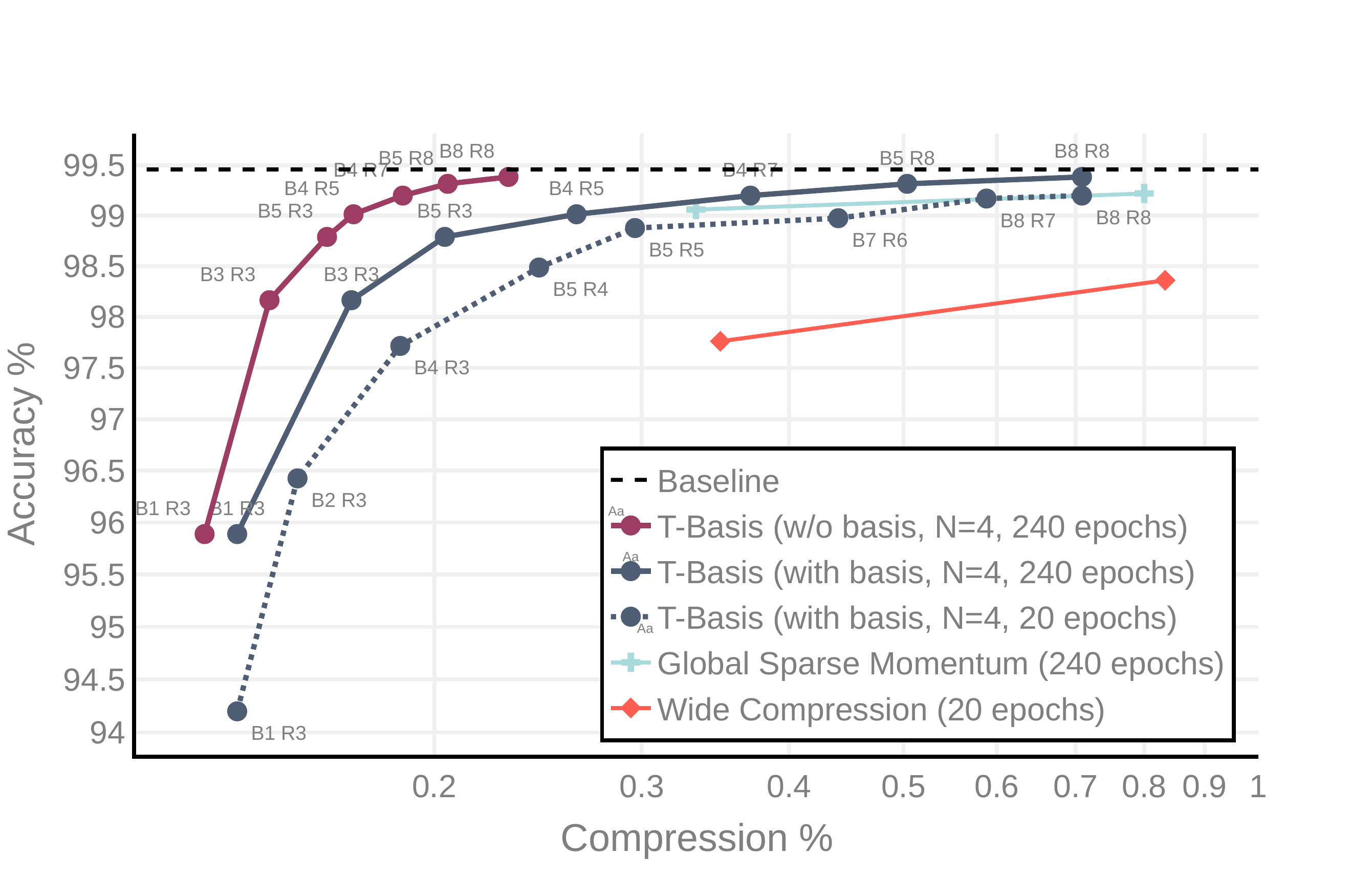}
\vspace{-0.6cm}
\caption{Top1 Accuracy vs. Model Compression Ratio for LeNet5 on MNIST classification. Each data point is annotated with basis size and rank. Our method outperforms in-place Tensor Ring weights allocation scheme and Global Sparse Momentum under extended training protocol. Legend: Wide Compression (TR)~\cite{wang2018wide}, Global Sparse Momentum SGD~\cite{ding2019global}}
\label{img:results_mnist}
\end{figure}

\subsection{A Note on the Training Time}

Training protocols of reference CNNs are designed with respect to convergence rates of the uncompressed models \cite{he2016deep, simonyan2014very}. Many works \cite{deeplabv3p} bootstrap existing feature extractors pre-trained on ImageNet~\cite{krizhevsky2012imagenet}, which helps them to reduce training time\footnote{\emph{Training time}, \emph{longer}, or \emph{shorter} terms are used in the context of the number of training iterations or epochs.} \cite{he2019rethinking}. 

In most neural network compression formulations, customized architectures cannot be easily initialized with pre-trained weights. Hence longer training time is required for compressed models to bridge the gap with reference models. The increases in training time seen in the prior art can be as high as 12$\times$ \cite{ding2019global}; others report results of training until convergence~\cite{eban2019structured}.

Following this practice, we report the performance of a few experiments under the extended training protocol and conclude that longer training time helps our models in the absence of a better than random initialization.

\subsection{Small Networks: LeNet5 on MNIST}

We follow the training protocol explained in \cite{wang2018wide}: 20 epochs, batch size 128, network architecture with two convolutional layers with 20 and 50 output channels respectively, and two linear layers with 320 and 10 output channels, a total of 429K in uncompressed parameters. Fig.~\ref{img:results_mnist} demonstrates performance of the specified LeNet5 architecture~\cite{lenet5} and \tbasis parameterization with various basis sizes and ranks. When training for 240 epochs, significant performance improvements are observed, and the model becomes competitive with~\cite{ding2019global}, using the same extended training protocol.

\paragraph{Ablation Study: Pseudo-random \tbasis}

\begin{figure}[t]
\centering
\includegraphics[trim={0 0cm 0 2cm},clip,width=1.05\linewidth]{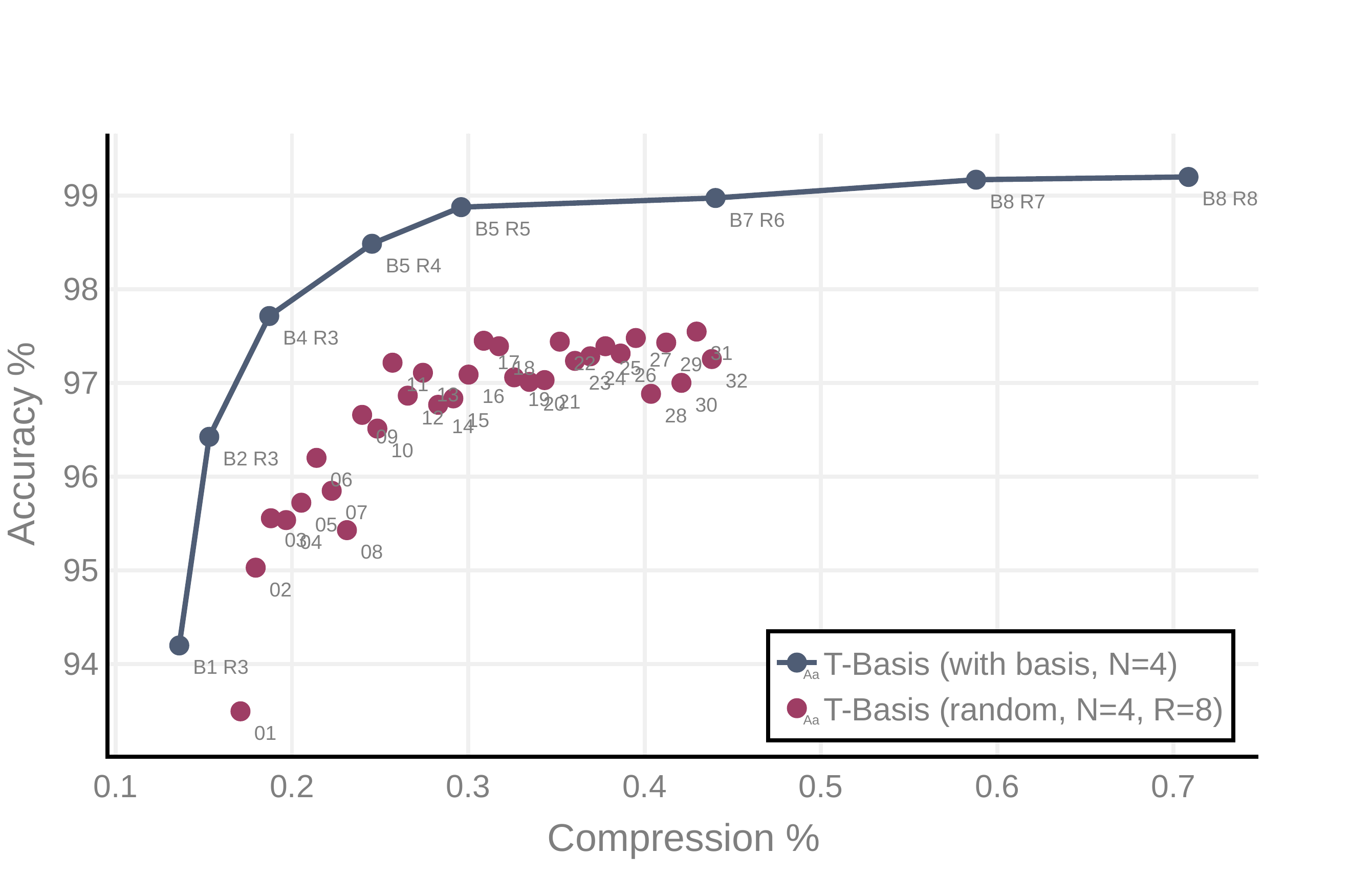}
\vspace{-0.6cm}
\caption{Comparison of learned and pseudo-random \tbasis approaches on MNIST digits classification with LeNet5. The performance of the learned basis cannot be reached by the pRNG basis, even with a large basis size (shown in datapoints annotations).}
\label{img:results_mnist_random}
\end{figure}

A natural question arises whether \tbasis basis needs to be a part of the learned parameters -- perhaps we could initialize it with a pseudo-random numbers generator (pRNG) with a fixed seed (the fourth integer value describing all elements of such \tbasis other than B, R, and N) and only learn $\bm{\alpha}$ and $\bm{\rho}$ parameters? Fig.~\ref{img:results_mnist_random} attempts to answer this question. In short, pRNG \tbasis approach is further away from the optimality frontier than its learned counterpart.

\subsection{Medium Networks: ResNets on CIFAR Datasets}
\begin{figure}[t]
\centering
\includegraphics[trim={0 0cm 0 2.0cm},clip,width=1.05\linewidth]{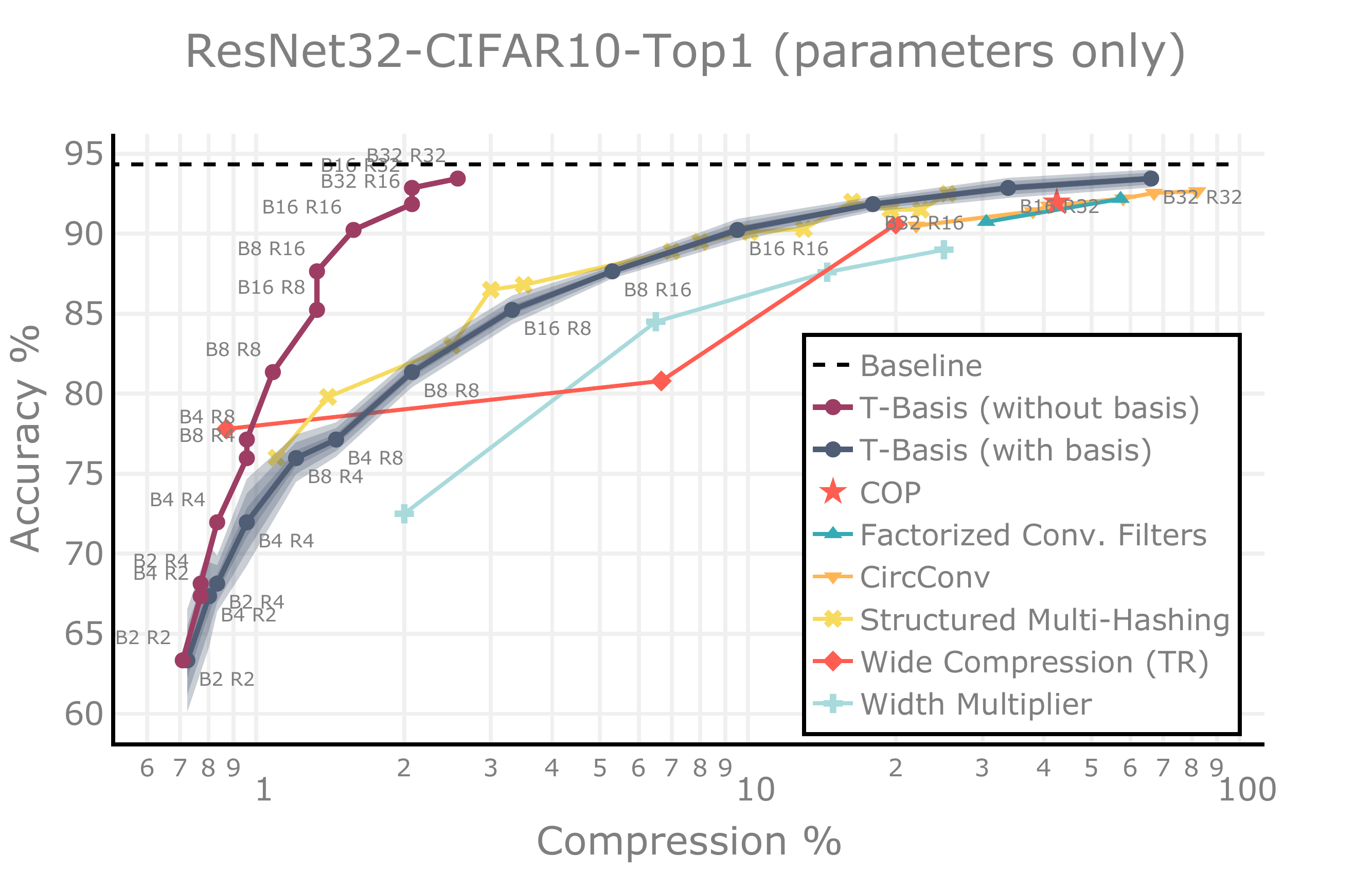}
\vspace{-0.6cm}
\caption{
Top1 Accuracy vs. Model Compression Ratio for ResNet-32 on CIFAR-10 image classification. Shaded areas correspond to confidence intervals for one, two, and three standard deviations, computed over 10 runs. Our method outperforms in-place Tensor Ring weights allocation scheme~\cite{wang2018wide}, and is on par or better than other state-of-the-art methods. Legend: COP~\cite{wang2019cop}, Factorized Conv. Filters~\cite{li2019compressing}, CircConv~\cite{liao2019circconv}, Structured Multi-Hashing~\cite{eban2019structured}, Wide Compression (TR)~\cite{wang2018wide}, Width Multiplier~\cite{eban2019structured}.
}
\label{img:results_cifar10}
\end{figure}
\begin{figure}[t]
\centering
\includegraphics[trim={0 0cm 0 2.0cm},clip,width=1.05\linewidth]{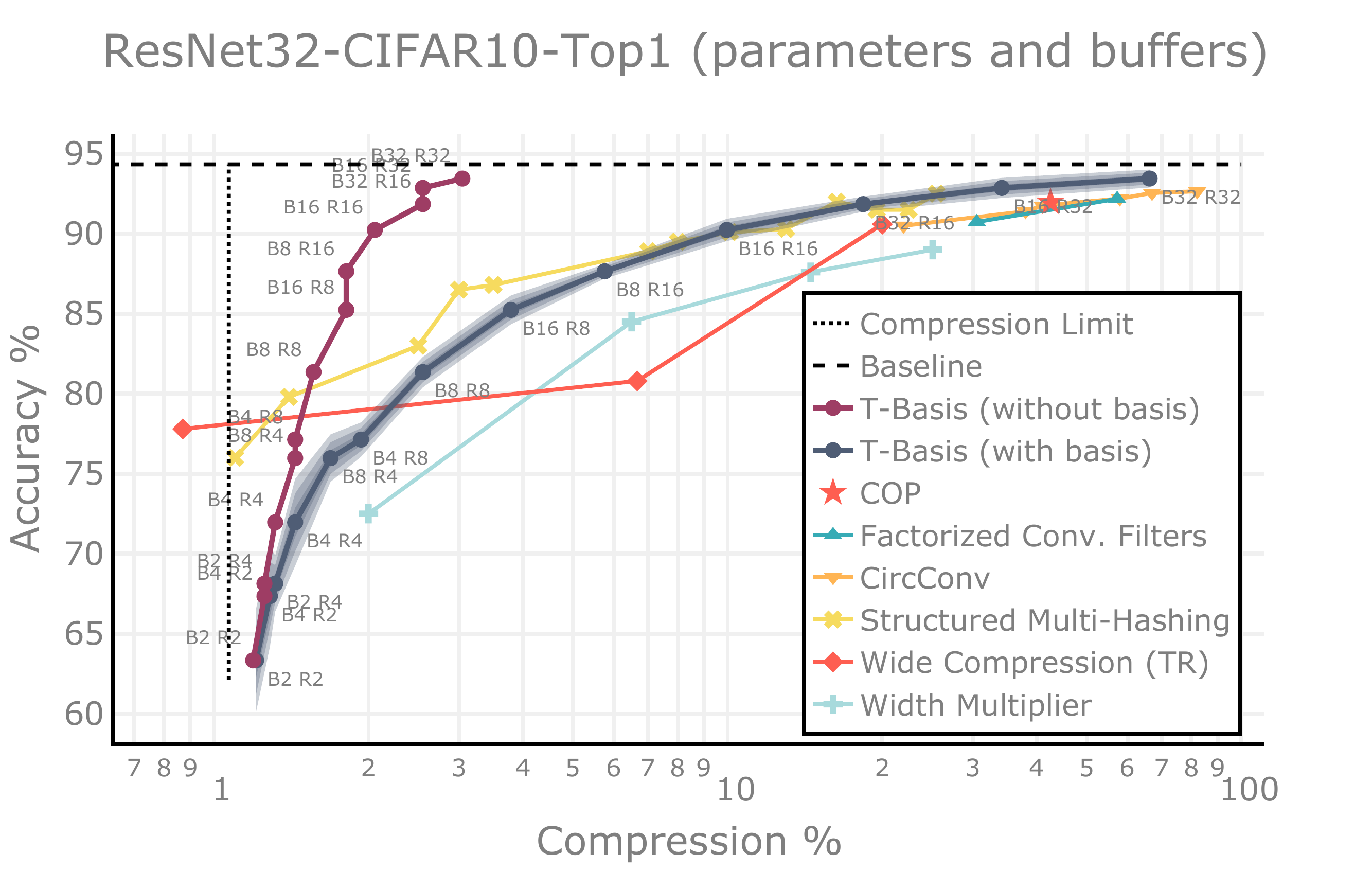}
\vspace{-0.6cm}
\caption{Top1 Accuracy vs. Model Compression Ratio for ResNet-32 on CIFAR-10 image classification. \textit{In this plot, \tbasis compression ratio is computed for compressible and incompressible parameters, and incompressible buffers, such as batchnorm statistics}. Compression Limit denotes the ratio of incompressible and original network sizes. Wide Compression (TR) reports one point with smaller ratio than the limit, and Structured Multihashing reports one experiment right on the limit, suggesting that at least both of these methods report conventional compression ratios as explained in Sec.~\ref{experiments}. The rest of the legend is the same as in Fig.~\ref{img:results_cifar10}.
}
\label{img:results_cifar10_total}
\end{figure}

CIFAR-10 \cite{cifar10} and ResNet-32 \cite{he2016deep} are arguably the most common combination seen in network compression works. This dataset consists of 60K RGB images of size 32x32, split into 50K train and 10K test splits. ResNet32 consists of a total of 0.46M parameters. Similarly to the prior art, we train our experiments until convergence (for 1000 epochs) with batch size 128, initial learning rate 0.1, and 50\%-75\% step LR schedule with gamma 0.1. The results are shown in Fig.~\ref{img:results_cifar10} for the standard compression ratio evaluation protocol, and in Fig.~\ref{img:results_cifar10_total} for the protocol which accounts for incompressible buffers.
As in the previous case, our method outperforms in-place Tensor Ring weights allocation scheme \cite{wang2018wide} and is on par or better than other state-of-the-art methods. 
It is worth noting that the line ``\tbasis w/o basis" contains vertical segments, corresponding to pairs of points with $(B,R)=(X,Y)$ and $(B,R)=(Y,X)$, with larger rank consistently giving a better score. 

\paragraph{Ablation Study: Rank Adapters}

Experimenting with different activation functions applied to the rank adapters values revealed that $\mathrm{exp}$ function yields the best results. No rank adaptation incurred $\sim$1\% drop of performance in experiments with ResNet-32 on CIFAR-10.

\begin{figure}[t]
\centering
\includegraphics[trim={0 0 0 2cm},clip,width=1.05\linewidth]{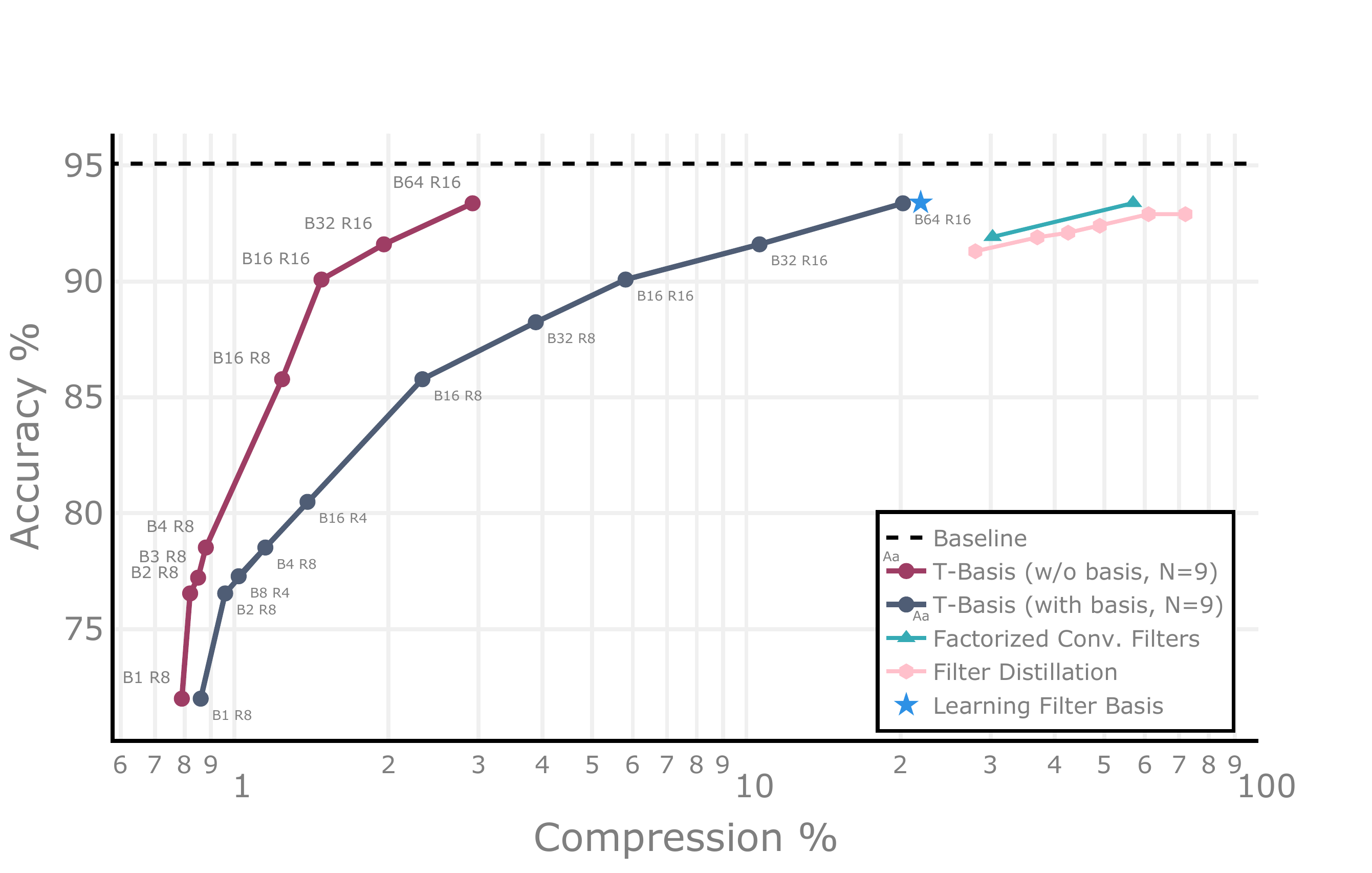}
\vspace{-0.6cm}
\caption{Top1 Accuracy vs. Model Compression Ratio for ResNet-56 on CIFAR-10 classification. Legend: Factorized Conv. Filters~\cite{li2019compressing}, Filter Distillation~\cite{suau2018network}, Learning Filter Basis~\cite{li2019learning}.}
\label{img:results_resnet56}
\end{figure}
\begin{figure}[t]
\centering
\includegraphics[trim={0 0 0 2cm},clip,width=1.05\linewidth]{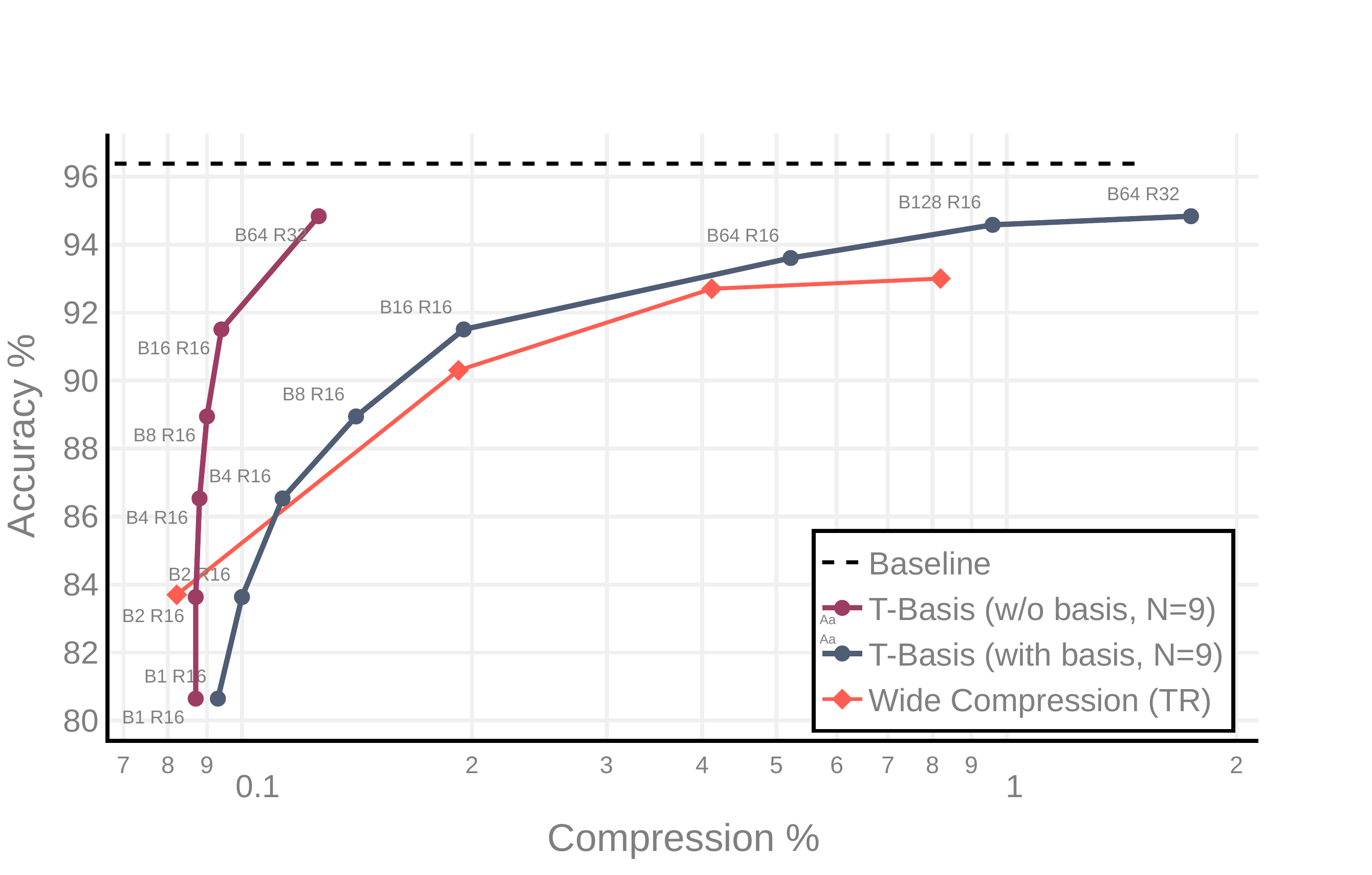}
\vspace{-0.6cm}
\caption{Top1 Accuracy vs. Model Compression Ratio for WRN-28-10 on CIFAR-10 classification. Legend: Wide Compression (TR)~\cite{wang2018wide}.}
\label{img:results_wrn2810}
\end{figure}
\begin{table}[t]
\vspace{-0.2cm}
\caption{Comparison of \tbasis with Tensor Ring Nets \cite{wang2018wide} compressing (w/ Basis) WRN-28-10 on CIFAR-100.}
\begin{center}
\begin{tabular}{lcc}
\toprule
 Method & Top1 Acc. & Compress. \% \\
 \midrule
 \tbasis ($B32$ $R8$) & $57.64$ & $\mathbf{0.149}$ \\
 \tbasis ($B32$ $R8$ w/o B.) & $57.64$ & $\mathbf{0.098}$ \\
 TR & $56.1$ & $0.239$\\
 Baseline & $78.3$ & $100$\\
 \bottomrule
\end{tabular}
\end{center}
\label{tab:wideresnet}
\end{table}

\paragraph{ResNet-56}
ResNet56 consists of a total of 0.85M uncompressed parameters. In order to compare with the results of \cite{li2019learning}, who also capitalize on the basis representation, we comply with their training protocol and train for 300 epochs. Results are presented in Fig.~\ref{img:results_resnet56}. It can be seen that \tbasis is on par with ``Learning Filter Basis'', and outperforms other state-of-the-art by a significant margin.

\paragraph{WideResNet-28-10}
WRN-28-10 is an enhanced ResNet architecture~\cite{wideresnet}; it contains 36.5M uncompressed parameters, which is considered a heavy-duty network. 
We compare with \cite{wang2018wide} in Table \ref{tab:wideresnet} and Fig.~\ref{img:results_wrn2810} in the extreme compression regime.

\subsection{Large Networks: DeepLabV3+ on Pascal VOC}

DeepLabV3+ \cite{deeplabv3p} is a standard network for semantic segmentation. We choose ResNet-34 backbone as a compromise between the model capacity and performance. Pascal VOC \cite{voc} and SBD \cite{sbd} datasets are often used together as a benchmark for semantic segmentation task. They are composed of 10582 photos in the training and 1449 photos in the validation splits and their dense semantic annotations with 21 classes. We train for 360K steps, original polynomial LR, batch size 16, crops 384$\times$384, without preloading ImageNet weights to make comparison with the baseline fair. 

\begin{table}
\caption{mIoU vs. Model Compression Ratio (w/ Basis) for DeepLabV3+ on VOC+SBD semantic segmentation. Confidence interval: $mean \rpm std$ over 3 runs. 
}
\begin{center}
\begin{tabular}{lcc}
\toprule
 Method & mIoU & Compress. \% \\
 \midrule
 \tbasis ($B128$ $R16$) & $58.18 \rpm 0.69$ & $1.450$ \\
 \tbasis ($B128$ $R32$) & $63.07 \rpm 0.08$ & $4.777$ \\
 Baseline & $64.08 \rpm 0.59$ & $100.0$ \\
 \bottomrule
\end{tabular}
\end{center}
\label{tab:deeplabv3p}
\end{table}

\begin{figure}[t]
\centering
\includegraphics[width=1\linewidth]{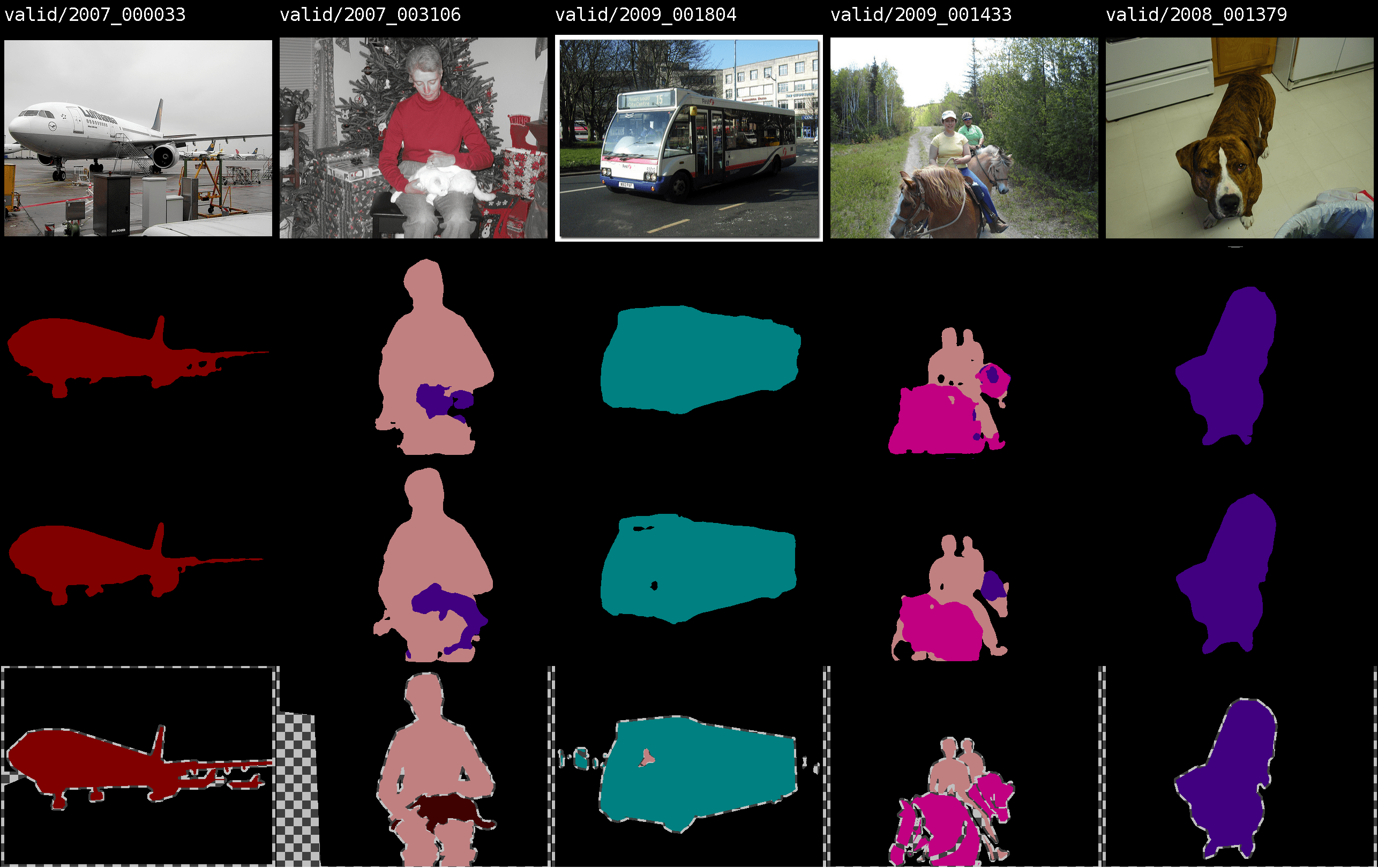}
\vspace{-0.2cm}
\caption{Semantic Segmentation predictions for a few images from Pascal VOC dataset validation split, made by a DeepLabV3+ model with ResNet-34 backbone, and 21$\times$ reduction in the number of parameters. Top to bottom: Input image, prediction by the compressed model, prediction by baseline, ground truth.}
\label{img:semseg}
\end{figure}

Our best results (see Table~\ref{tab:deeplabv3p}) for semantic segmentation are 1\% away from the baseline with $<$5\% of weights. A few visualizations from the validation split can be found in Fig.~\ref{img:semseg}. We conclude that transitioning from toy datasets and models to large ones requires ranks $\geq 32$ in order to maintain performance close to the baseline.

\section{Conclusion}
\label{sec:conclusion}

We introduced a novel concept for compressing neural networks through a compact representation termed \tbasis. 
Motivated by the uniform weights sharing criterion and the low-rank Tensor Ring decomposition, our parameterization allows for efficient representation of Neural Networks weights through a shared basis and a few layer-specific coefficients. 
We demonstrate that our method can be used on top of the existing neural network architectures, such as ResNets, and introduces just two global hyperparameters -- basis size and rank.
Finally, \tbasis parameterization supports a broad range of compression ratios and provides a new degree of freedom to transfer the basis to other networks and tasks. 
We further study low-rank weight matrix parameterizations in the context of neural network training stability in our follow-up work~\cite{obukhov2021spectral}.

\section*{Acknowledgements}
\label{sec:ack}

This work is funded by Toyota Motor Europe via the research project TRACE-Zurich. We thank NVIDIA for GPU donations, and Amazon Activate for EC2 credits. Computations were also done on the Leonhard cluster at ETH Zurich; special thanks to Andreas Lugmayr for making it happen. 
We thank our reviewers and ICML organizers for their feedback, time, and support during the COVID pandemic.

{
\balance
\bibliography{main}
\bibliographystyle{icml2020}
}

\end{document}